\documentclass[a4paper]{article}
\usepackage{fullpage}
\usepackage{authblk}

\usepackage[utf8]{inputenc} 
\usepackage[T1]{fontenc}    
\usepackage{hyperref}   
\usepackage{url}            
\usepackage{booktabs}      
\usepackage{amsfonts}      
\usepackage{nicefrac}      
\usepackage{microtype}     
\usepackage{graphicx} 
\usepackage{bm,dsfont}
\usepackage{amssymb,amsmath,amsthm,mathtools}
\usepackage{multirow}
\usepackage{subcaption}
\usepackage{bbm}

\def\eqref#1{equation~\ref{#1}}

\def\1{\bm{1}}

\def\rvd{{\mathbf{d}}}

\def\rvs{{\mathbf{s}}}

\def\rvx{{\mathbf{x}}}

\def\rvz{{\mathbf{z}}}

\DeclareMathAlphabet{\mathsfit}{\encodingdefault}{\sfdefault}{m}{sl}
\SetMathAlphabet{\mathsfit}{bold}{\encodingdefault}{\sfdefault}{bx}{n}

\DeclareMathOperator*{\Sim}{sim}

\title{Self-Supervised Cross-Encoder for
Neurodegenerative Disease Diagnosis}

\author[1]{Fangqi Cheng}
\author[2]{Yingying Zhao}
\author[1]{Xiaochen Yang\footnote{Equal contribution}\footnote{Corresponding author; Email address: xiaochen.yang@glasgow.ac.uk}}
\affil[1]{School of Mathematics and Statistics, Glasgow, UK.}
\affil[2]{Department of Computer and Information Sciences, University of Strathclyde, Glasgow, UK.}
\date{}

\begin{document}

\maketitle

\begin{abstract}
Deep learning has shown significant potential in diagnosing neurodegenerative diseases from MRI data. However, most existing methods rely heavily on large volumes of labeled data and often yield representations that lack interpretability. To address both challenges, we propose a novel self-supervised cross-encoder framework that leverages the temporal continuity in longitudinal MRI scans for supervision. This framework disentangles learned representations into two components: a static representation, constrained by contrastive learning, which captures stable anatomical features; and a dynamic representation, guided by input-gradient regularization, which reflects temporal changes and can be effectively fine-tuned for downstream classification tasks. Experimental results on the Alzheimer’s Disease Neuroimaging Initiative (ADNI) dataset demonstrate that our method achieves superior classification accuracy and improved interpretability. Furthermore, the learned representations exhibit strong zero-shot generalization on the Open Access Series of Imaging Studies (OASIS) dataset and cross-task generalization on the Parkinson Progression Marker Initiative (PPMI) dataset. The code for the proposed method will be made publicly available.
\end{abstract}

\section{Introduction}
\label{sec1}
With the development of medicine, science, and technology, people's life expectancy is increasing nowadays, consequently leading to a rise in the number of people diagnosed with neurodegenerative diseases, such as Alzheimer's disease~(AD) and Parkinson's disease~(PD)~\cite{hou2019ageing}. In clinical practice, neurologists often use structural magnetic resonance imaging~(MRI) scans to assist diagnosis~\cite{frisoni2010clinical}. However, interpreting MRI scans demands a high level of medical expertise. In recent years, there has been growing research on developing machine learning methods for classifying neurodegenerative diseases based on MRI data~\cite{noor2020application,saikia2024alzheimer}. These methods can enhance the efficiency and scalability of diagnosis and have the potential of improving diagnosis accuracy, especially in the early stage of the disease where subtle changes may be overlooked~\cite{singh2023early, ghaffar2023evaluation}.  

While showing promise, machine learning methods, particularly deep learning methods, face the challenge of insufficient MRI data with accurate labels to effectively train the models~\cite{tajbakhsh2016convolutional}. Recent studies on deep learning suggest that the label scarcity issue can be addressed through self-supervised learning, which uses the data itself as the supervision information to learn and extract the useful features~\cite{liu2021self, jing2020self}. Compared with natural images which often require deliberate design of pretext tasks to generate supervision, longitudinal MRI data, acquired from the same subjects over multiple visits~\cite{cascarano2023machine}, contains temporal information that naturally serves as the supervisory signal, aiding in the identification of features that change over time. In addition to the above limitation, deep neural networks tend to generate feature representations laden with redundant information and lacking interpretability~\cite{alzubaidi2021review}. While considerable effort have been made, most interpretability approaches are post-hoc, meaning that they provide insights into the decision-making process after training the model, rather than being inherently integrated into model~\cite{martin2023interpretable,vimbi2024interpreting}. The reason is that intrinsically interpretable methods often cannot reach a very ideal performance~\cite{kang2023interpretable}.

To address the aforementioned challenges, this paper proposes a self-supervised cross-encoder framework that learns disentangled feature representations from unlabeled MRI scans, which can be easily fine-tuned for classifying neurodegenerative diseases while possessing interpretability. Specifically, we adapt the cross-encoder, originally proposed in~\cite{sun2021cross} for gaze estimation, to model structural and disease-related aspects of the brain. In contrast to classical autoencoders which encode a single image into a latent vector and then decode it to reconstruct the same image, our cross-encoder takes as input a pair of brain images from the same subject acquired at two different time points and reconstructs both images simultaneously. More importantly, the model splits the latent vector into two parts and swaps one part between the image pair before reconstruction. Such design enables the model to learn two types of features: the swapped part captures \emph{static features} that represent time-invariant aspects of brain structure, while the non-swapped part captures \emph{dynamic features} that reflect temporal changes in brain, likely related to aging or brain diseases. 

Moreover, to ensure that the static features are informative, we employ contrastive learning~\cite{chen2020simple, ccaugatan2024sigclr} to encourage static features of the same subject to be similar and those of different subjects to be dissimilar. Furthermore, we apply an $L_1$-norm regularization to the gradient of dynamic features with respect to their inputs, which consequently selects and emphasizes a small subset of brain voxels that are responsible for time-varying changes in the brain. The proposed method is evaluated on the Alzheimer’s Disease Neuroimaging Initiative (ADNI) dataset~\cite{ADNI}, and results suggest that the learned representations improve AD classification, whether they are frozen or fine-tuned, and the discriminative regions identified by the model align better with the brain regions associated with AD. In addition, the learned representations remain effective when evaluated on an unseen AD dataset -- the Open Access Series of Imaging Studies (OASIS) -- and on a different neurodegenerative disorder, Parkinson's disease~(PD), demonstrating the strong generalization ability of the proposed method across both same-domain datasets and related tasks.

The contribution of this work can be summarized as follows:
\begin{itemize}
\item We propose a self-supervised cross-encoder network, which by forcing the model to reconstruct images using partially swapped representations, effectively disentangles dynamic, potentially disease-related features from static features using unlabeled MRI scans.
\item Contrastive learning and input-gradient regularization are utilized to enforce the static and dynamic features to be more informative, as well as further improving the model interpretability.
\item Quantitative and visualization results on the AD and PD datasets demonstrate the superior performance of our method over the state-of-art methods in terms of classification accuracy, interpretability, and generalizability, with ablation studies further showing the effectiveness of the aforementioned two constraints in enhancing the feature representations. 
\end{itemize}

\section{Related Work}
\label{sec2}


\subsection{Methods of Neurodegenerative Disease Classification}

Neurodegenerative classification often relies on various bio-markers. Some studies explore the integration of neuroimaging data and cognitive scores to model disease progression~\cite{gou2024tensor}. Notably, even relying solely on neuroimaging modalities proves effective in such tasks, with MRI being one of the most widely adopted~\cite{ADtanveer2020machine, PDReview}. For instance, according to Tanveer et al.'s review~\cite{ADtanveer2020machine}, approximately 40\% of studies utilizing neuroimages for AD classification opt for MRI. According to Jie et al.~\cite{PDReview}, around 11\% of studies use MRI images for PD classification.

Many existing works show that deep learning methods are effective in learning discriminative feature representations from raw MRI data and can accurately classify neurodegenerative diseases when given a large amount number of neuroimages~\cite{tuauctan2021artificial}. However, considering the high cost of annotating the data, it is imperative to develop deep learning methods that work well on limited labeled data. There are several ways to deal with the lack of labeled data. For example, transfer learning methods first pre-train a model on a large dataset, which may consist of natural images~\cite{ashraf2021deep} or neuroimages irrelevant to the disease of interest~\cite{cai2022graph}, and then fine-tune the model and/or learn a simple classifier on the downstream dataset. Cai et al.~\cite{cai2022graph} pre-trains an autoencoder on the UK Biobank dataset to learn the latent representations of structural MRI and diffusion tensor imaging data. Following this, a graph transformer network is trained to learn the difference between the estimated brain age and chronological age, an important biomarker of AD diagnosis. The learned feature representations and estimated differences are  finally used as the input for the AD classifier. 
Other approaches to addressing limited labeled data include semi-supervised learning, which leverages a small amount of labeled data and a large amount of unlabeled data for training~\cite{kingma2014semi}, and weakly supervised learning, where Ouyang et al.~\cite{ouyang2022IEEE}  orthogonally encode aging trajectories and disease severity by leveraging samples from different diagnostic groups. Semi-supervised learning and weakly supervised learning typically still require some labeled data or supervisory signals to guide the learning process~\cite{zhao2023comparison}. 

More recently, research has emphasized learning representations that capture local brain structure for AD diagnosis. For example, Wang et al.~\cite{wang2026ah} utilize contrastive learning-based pre-training to capture both local and global brain structural information. Despite these advances, existing methods still struggle to learn disease-related local representations effectively when only limited labeled data are available.

\subsection{Self-supervised Learning in Neurodegenerative Disease}

Self-supervised learning~(SSL) can train models entirely on unlabeled data~\cite{liu2021self}. Generative SSL methods learn representations from pretext tasks, such as colorization~\cite{zhang2016colorful} and rotation~\cite{gidaris2018unsupervised}. Contrastive SSL methods learn representations by maximizing similarity between similar instances and minimizing it between dissimilar ones~\cite{chen2020simple}.

Recent SSL studies for neurodegenerative diseases use longitudinal data to capture variable progression. For example,~\cite{zhao2021longitudinal} aligns representations of MRI pairs with a linear direction encoding brain aging. \cite{ouyang2022self} models inter-subject trajectory vector similarity using non-linear directions in the trajectory field. However, the representations learned by these methods capture information that is not related to time or disease, resulting in redundant representations which impose a burden when fine-tuning for downstream tasks. \cite{couronne2021longitudinal} disentangles the global disease timeline using medical longitudinal data~(e.g., cognitive scores and MRIs) by assuming disease progression factor and individual variability factor are independent from each other, constructing a model for disease progression. This approach effectively utilizes clinical data and demonstrates that a single direction in the latent space can capture temporal progression. More recently, \cite{liu2025simcmc} proposes using multi-view brain MRI slices and maximizes correlation between views to extract more compact representations. 

Notably, neither these nor prior methods inherently produce interpretable models or effectively extract disease-specific information. To address this limitation, we propose the use of cross-encoders, which disentangle disease-related from brain-structure-related features by modeling paired inputs. The original cross-encoder~\cite{sun2021cross} requires two types of paired inputs to learn two disentangled features. However, the AD and PD datasets in our study provide only one type of paired input, sufficient for learning dynamic features only. To compensate for the absence of the other paired input needed for static features, we incorporate contrastive learning. Together, our method explicitly learns compact, disease-specific temporal representations in an interpretable manner while avoiding redundant features.

\subsection{Interpretability}

While achieving high accuracy in downstream tasks is important, it is also crucial to identify the reasoning behind the classification decisions. However, most current models rely on post-hoc interpretation using external interpretability methods, such as Grad-CAM~\cite{songimage-gradcam}, CNN saliency maps~\cite{bron2021cross-CNNmap}, and guided-backpropagation~\cite{backpropagation}. These external methods are considered to lack uniqueness~\cite{nguyen2023towards}. \cite{yee2021construction} uses two different explainable methods in its work but gets dissimilar effects. \cite{zhu2022deep} proposes a linear multi-modal knowledge distillation module which learns a single-layer linear mapping to approximate the feature representations obtained from a deep encoder. The obtained feature can be used to reflect the importance of brain regions. 
While the linear layer is optimized jointly with the deep encoder, this method can still be considered post-hoc, as the training of the linear layer does not affect the deep representation learning. Regarding intrinsic interpretable methods, \cite{kang2023interpretable} indicates that these methods often do not exhibit very good accuracy, as they tend to trade-off between accuracy and interpretability. 
Therefore, it is desirable to design a novel interpretable neural network for neurodegenerative diseases without performance loss. Recent studies, such as ~\cite{zheng2025structured}, leverage structured domain knowledge combined with deep alignment to build interpretable models. While effective in incorporating prior information, these methods still cannot fully disentangle disease-specific representations or produce inherently interpretable features. In this work, we explicitly integrate interpretability into the model by combining input-gradient regularization with the disentanglement of static and dynamic brain features. This unified design enables the learning of compact, disease-specific representations that are both accurate and intrinsically interpretable, effectively addressing the key challenges outlined above. While input-gradient regularization has previously been explored to enhance adversarial robustness in neural networks~\cite{ross2018improving}, its application as a mechanism for learning interpretable features represents a novel contribution of this study.


\section{Method}
\label{sec3}

\begin{figure*}[tbp]
\centering
\includegraphics[width=0.8\textwidth]{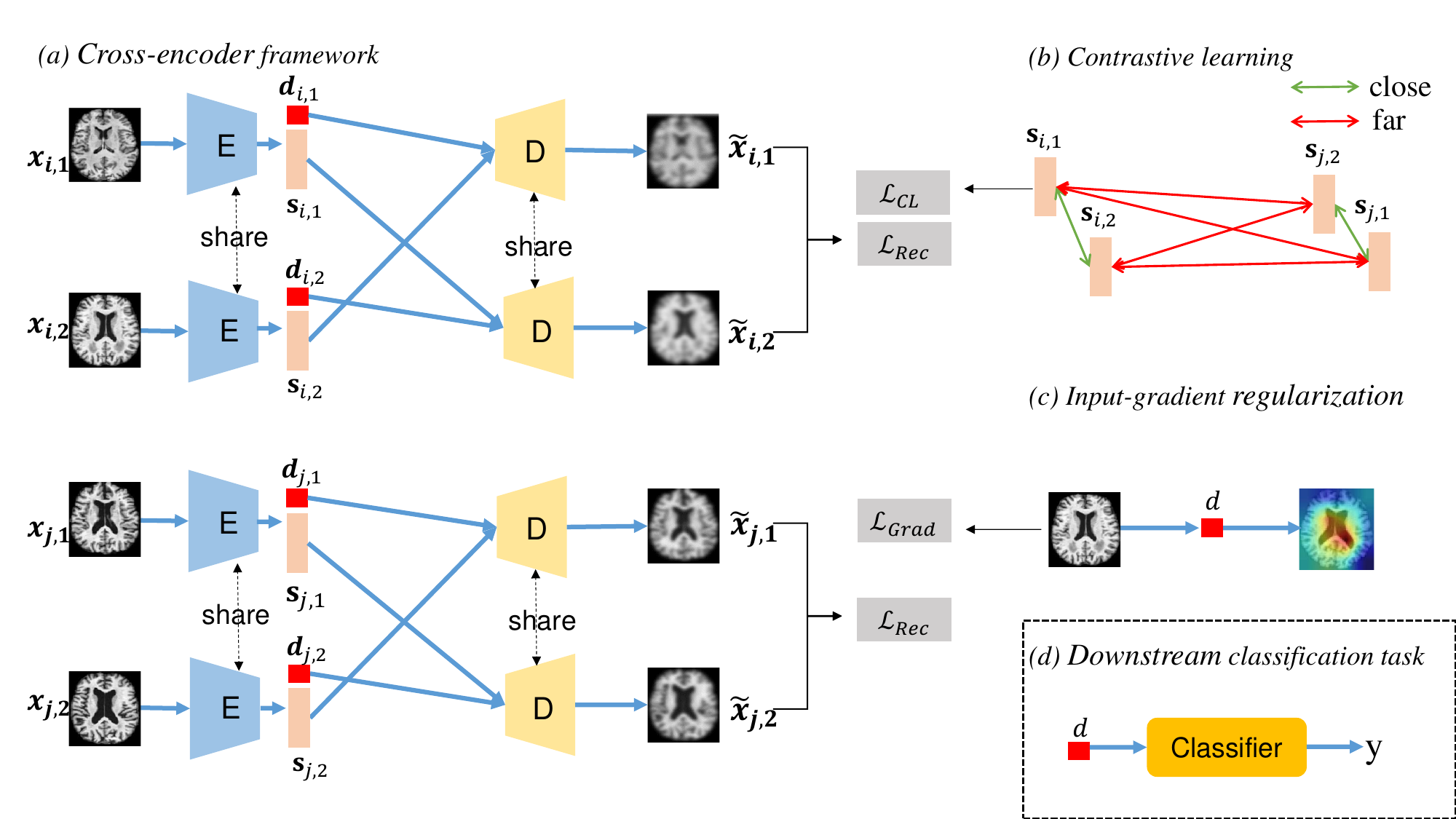}
\caption{\label{fig:Framework}Overview of the proposed framework. \textit{Fig.~(a)} presents the cross-encoder architecture. The encoder $E$ encodes an image into disentangled representations, consisting of dynamic features ($\rvd$) and static features ($\rvs$)~(the dimension of dynamic feature $\rvd$ is much smaller than $\rvs$). After swapping $\rvs$, the decoder $D$ reconstructs the images.  In $x_{i/j,1/2}$, $i$ and $j$ denote the different subjects, and $1$ and $2$ denote the different time points. The static features are further enhanced with contrastive learning (\textit{Fig.~(b)}), and dynamic features are constrained by input-gradient regularization (\textit{Fig.~(c)}). Once self-supervised training is complete, the dynamic features $\rvd$ will be used in the downstream classification task (\textit{Fig.~(d)}).}
\end{figure*}

As diseases progress, some brain structures change substantially in the same subject, while others remain relatively stable. We define features as dynamic and static if they correspond to time-varying part or the unchanging part, respectively. To learn such features, we combine the cross-encoder framework with contrastive loss and input-gradient regularization. The cross-encoder is primarily responsible for learning disentangled representations, while the contrastive loss and input-gradient regularization improve the informativeness of static features and dynamic features respectively. The whole framework is shown in Figure~\ref{fig:Framework}.

\subsection{Cross-Encoder Architecture}

To achieve feature disentanglement, the cross-encoder is developed based on the hypothesis that the cognitive decline affects only a small portion of the brain, while the remaining structures remain largely stable over time. Consequently, in the latent space of image features, the feature vectors can be partitioned into a small set of dynamic features and a larger set of static features.

\begin{figure*}[tbp]
\centering
\includegraphics[width=0.9\textwidth]{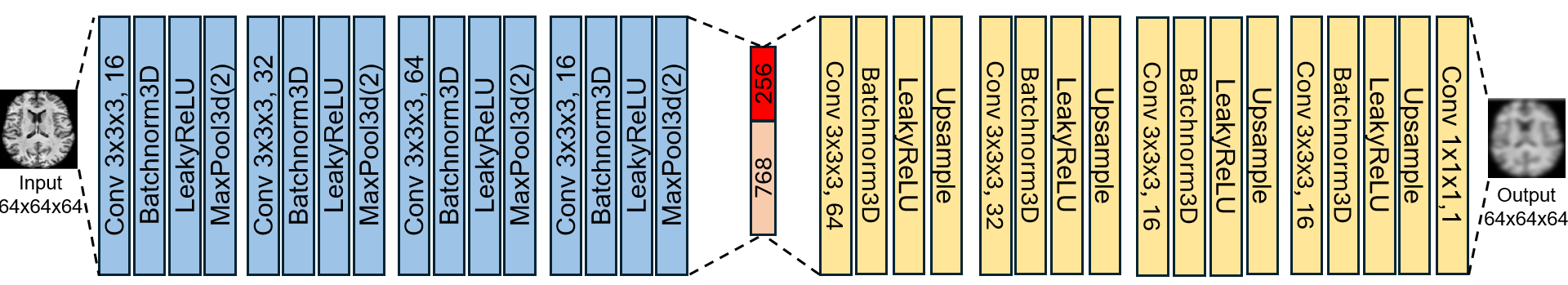}
\caption{Architecture of the cross-encoder network. The blue blocks are the encoder and the yellow blocks are the decoder. The encoder encodes the input MRI into a 1024-dimension latent vector, where dynamic features account for 25\% (red block) and static feature account for 75\% (orange block). The proportion of dynamic features is set as a hyperparameter and is evaluated through sensitivity analysis in later experiments.}
\label{fig:encoder-decoder}
\end{figure*}

As shown in Figure~\ref{fig:encoder-decoder}, the proposed framework comprises an encoder and a decoder. The encoder extracts latent representations from MRI scans and disentangles them into \textit{dynamic features} related to normal aging and cognitive decline and \textit{static features} largely unrelated to disease progression. Specifically, the encoder consists of four blocks, each of which contains a convolutional layer, batch normalization, LeakyReLU activation, and max pooling. The decoder also consists of four blocks, each containing a convolutional layer, batch normalization, LeakyReLU activation, and upsampling, followed by an extra convolutional layer for ultimate reconstruction. 

To train the cross-encoder, we only use MRI images from subjects with at least two visits. From these subjects, we can construct a set of image pairs $\{(\rvx_{i,1},\rvx_{i,2})\}_{i=1}^{|\mathcal{I}|}$, and $\mathcal{I}$ denotes the set of all subjects\footnote{A subject with $n$ visits will generate $n \choose 2$ image pairs. For example, if a subject contains three images $\rvx_1,\rvx_2,\rvx_3$, then three image pairs will be obtained, namely $(\rvx_1, \rvx_2), (\rvx_1, \rvx_3), (\rvx_2, \rvx_3)$. In the training phase, each batch will only include one image pair from each subject.}. The cross-encoder network, shown in the upper panel of Figure~\ref{fig:Framework}(a), takes a pair of MRI images ($\rvx_{i,1}, \rvx_{i,2}$) from the same subject at different time points as input. By using an encoder $E$, the inputs are transformed into latent vectors $\rvz_{i,1}$ and $\rvz_{i,2}$. To achieve interpretabiliy, we forcefully split the latent vectors into two parts: $\rvz_{i,1}=[\rvd_{i,1},\rvs_{i,1}]$ and $\rvz_{i,2}=[\rvd_{i,2},\rvs_{i,2}]$, where $\rvd_{i,\cdot}$ are termed dynamic features and $\rvs_{i,\cdot}$ are termed static features. Next, we swap the position of $\rvs_{i,\cdot}$ in the pair and use a decoder $D$ to generate the reconstructed image $\tilde{\rvx}_{i,1}$ from $[\rvd_{i,1},\rvs_{i,2}]$ and $\tilde{\rvx}_{i,2}$ from $[\rvd_{i,2},\rvs_{i,1}]$. To ensure that the reconstructed images closely resemble the original images, the following reconstruction loss is used: 
\begin{equation}\begin{split}
\mathcal{L}_\text{Rec} &= \sum_{i\in \mathcal{I}} {\|\rvx_{i,1} - \tilde{\rvx}_{i,1}\|}_1 + {\|\rvx_{i,2}-\tilde{\rvx}_{i,2}\|}_1
\\
&=\sum_{i\in \mathcal{I}} {\|\rvx_{i,1} - D(\rvd_{i,1},\rvs_{i,2})\|}_1 + {\|\rvx_{i,2}-D(\rvd_{i,2},\rvs_{i,1})\|}_1.
\end{split}
\label{eq:cross_encoder}
\end{equation}

Before training the cross-encoder, the latent vector $\rvz$ is simply a randomized feature vector and thus the dynamic feature $\rvd$ and the static feature $\rvs$ are indistinguishable. However, as training progresses, our design of splitting $\rvz$ into two components and swapping $\rvs$ forces the static features to learn the commonality between input pairs and the dynamic features to learn their differences. In the context of neuroimaging data, dynamic features are expected to reflect changes in the brain, possibly due to aging or neurodegenerative diseases, and static features should be the same in one subject's brain across different time points. Since the static features are supposed to be time-invariant, swapping their positions will not affect the reconstruction of images. 

However, if we solely focus on minimizing the reconstruction error, an extreme situation might occur: to ensure no information is lost after swapping the static features, the dynamic features may contain all information about the images while the static features contain none. 
To avoid this issue, we set the dimension of $\rvd$ to be only a small proportion of the entire latent dimension. In addition, we utilize contrastive learning to constrain the static features and enforce input-gradient regularization to constrain the dynamic features as described below.

\subsection{Contrastive Loss}

We address the non-informativeness of static features by examining their similarities within and between subjects. In principle, apart from changes related to the disease, the brain structures of the same subject should remain relatively stable. Therefore,  the static features of the same subject at different time points should be similar. In contrast, due to inter-subject variability, static features of different subjects should exhibit differences. 

Such intuition can be formulated by employing contrastive learning~(CL). As illustrated in Figure~\ref{fig:Framework}(b), static features from subject $i$ ($\rvs_{i,1}$ and $\rvs_{i,2}$) should have a small distance, whereas their distances to static features from subject $j$ ($\rvs_{j,1}$ and $\rvs_{j,2}$) should be large. To implement CL, we adopt SigCLR~\cite{ccaugatan2024sigclr}. 
This approach explicitly promotes the clustering of representations from the same subject while enforcing a clear separation between different subjects. The SigCLR loss is defined by applying the logistic loss to pairs of static features:

\begin{equation}
\mathcal{L}_\text{CL} = 
-\frac{1}{|\mathcal{I}|} \sum_{i\in\mathcal{I}} \sum_{j\in\mathcal{I}}\sum_{k=1}^2\sum_{l=1}^2\mathbbm{1}_{(i,k) \ne (j,l)}
\log\left( \frac{1}{1 + \exp\left(  -z_{ij} \cdot(\tau\Sim(\rvs_{i,k},\rvs_{j,l}) +b) \right) } \right),
\label{eq:contrastive loss}
\end{equation}
where $\Sim$ denotes the cosine similarity, $\tau$ is a fixed temperature hyperparameter which controls the scaling of similarity scores, $b$ is a learnable bias which prevents loss from being almost completely dominated by negative samples. $z_{ij}=1$ if $i = j$ and $-1$ otherwise. $\mathbbm{1}_{(i,k) \neq (j,l)}$ is an indicator function that equals to 1 if $(i,k)\neq(j,l)$ and 0 otherwise, to exclude trivial self-pairs from the loss computation.

\subsection{Input-Gradient Regularization}

According to existing research on neuroscience, normal aging often results in a decrease in brain volume, particularly in the frontal lobe and hippocampus, an enlargement in ventricles, and wider and shallower sulci~\cite{lee2022normal}. In addition, individuals with AD often exhibit more pronounced changes in regions such as hippocampus~\cite{jack1997medial}, while those with PD show changes in substantia nigra. These studies inspire us to adopt the input-gradient regularization to focus attention on specific brain regions rather than the whole brain for detecting aging-related changes and neurological diseases. 

More specifically, we first calculate the gradient of the dynamic feature $\rvd$ with respect to the input image $\rvx$. This gradient quantifies the impact of each voxel in the input image on the dynamic features. Secondly, to enforce that only a fraction of the input image affects the dynamic features, an $L_1$-norm penalty is applied on the input gradient, which encourages sparsity in the input and thereby highlights only the most important voxels. The loss function is given as: 
\begin{equation}
\mathcal{L}_\text{Grad} 
= \sum_{i\in\mathcal{I}} \sum_{k=1}^2{\left\|\frac{\partial \rvd_{i,k}}{\partial \rvx_{i,k}} \right\|}_1.
\label{eq:gradients}
\end{equation}

\subsection{Training Strategy}

The overall loss function is obtained by combining Eqs.~(\ref{eq:cross_encoder}), (\ref{eq:contrastive loss}), (\ref{eq:gradients}):

\begin{equation}
\mathcal{L}=\mathcal{L}_\text{Rec}+\lambda_1\mathcal{L}_\text{CL} + \lambda_2\mathcal{L}_\text{Grad} ,
\label{eq:overall}
\end{equation}
where $\lambda_1$ and $\lambda_2$ denote the trade-off parameters. By minimizing this loss function, the cross-encoder is trained to effectively disentangle static features and dynamic features. The dynamic features $\rvd$, despite having a relatively low dimension, retain the essential information required for classification. Therefore, once the training of cross-encoder is complete, only $\rvd$ will be used for downstream classification tasks, with a classifier trained on top of $\rvd$.
In addition, the low-dimensionality nature of $\rvd$ reduces the risk of overfitting when training the classifier and/or fine-tuning $\rvd$.

\section{Experiments}
\label{sctn::exp}
In this section, the experimental setup is first described, including the datasets, implementation details, evaluation metrics, and baseline methods. We then present the overall classification performance of the proposed method compared with the state-of-the-art baseline methods, followed by an analysis of its ability to capture key domain features, thereby demonstrating its interpretability. Next, the generalizability of the learned representations is evaluated through zero-shot performance on an unseen dataset and cross-task performance on an entirely different task. Subsequently, ablation studies are performed to quantify the contribution of each framework component. Finally, a sensitivity analysis is performed to provide insights into the choice of key parameters and examine the effectiveness of disentangling dynamic features from static ones.

\subsection{Experimental Setup} 
\subsubsection{Datasets}
We use three publicly available neuroimaging databases for pre-training and downstream evaluation of the proposed cross-encoder framework. The model is first pre-trained on the Alzheimer's Disease Neuroimaging Initiative (ADNI) dataset~\cite{ADNI}, which contains 1448 longitudinal T1-weighted MRI images from 341 subjects, each with multiple time points. The model processes MRI data at the voxel level, enabling it to learn detailed spatial features directly from the images.  

For the downstream AD classification, we use the ADNI dataset to evaluate two tasks: (1) normal control~(NC) vs. AD, using 68/7/19 NC and 105/12/28 AD subjects for training/validation/testing, respectively (randomly split); (2) stable mild cognitive impairment (sMCI) vs. progressive mild cognitive impairment (pMCI), using 52/8/15 sMCI and 94/13/27 pMCI subjects for training/validation/testing, respectively. The first task assesses the model’s ability to diagnose diseases, while the second task evaluates its capacity to capture subtle changes associated with disease progression. 

Zero-shot generalization is tested using the Open Access Series of Imaging Studies (OASIS)-2 dataset~\cite{marcusoasis}, with 160 MRI scans from 30 AD and 30 NC subjects. Cross-task generalization is assessed on the Parkinson’s Progression Markers Initiative (PPMI) dataset~\cite{PPMI}, containing 451 MRI scans from 203 subjects, split into 37/5/10 NC and 113/13/25 PD subjects for training/validation/testing.

Table~\ref{tab:dataset} shows the demographic details of the three datasets. All MRI scans were pre-processed with a standard pipeline involving denoising, bias field correction, skull stripping, affine registration to a template, re-scaling to a 64 × 64 × 64 volume, and transforming image intensities to z-scores~\cite{ouyang2022IEEE,ouyang2021self}.
\begin{table}[ht]
\caption{Demographic summary of subjects from Alzheimer’s disease (AD) and Parkinson’s disease (PD) datasets used for model pre-training and downstream classification tasks.}
\label{tab:dataset}
\centering
\begin{tabular}{lllll}
\toprule
Task & Dataset & \# Subjects & Age range~(years) & \# Male/Female \\
\midrule
Pre-traing & ADNI & 341 & 54 - 91 & 179/162 \\
Classification~(AD vs. NC) & ADNI & 239 & 57 - 91 & 116/123 \\
Classification~(sMCI vs. pMCI) & ADNI & 209 & 56 - 89 & 85/124 \\
Classification~(AD vs. NC) & OASIS & 60 & 60 - 93 & 23/37 \\
Classification~(PD vc. NC) & PPMI & 203 & 31 - 82 &138/65 \\
\bottomrule
\end{tabular}
\end{table}

\subsubsection{Implementation Details}
To improve model generalizability, data augmentation is applied on-the-fly during both the training and downstream classification stages. The augmentations include small random rotations (up to 4 degrees), shifts (within 4 voxels), and random flips along a spatial axis. These transformations help the model become invariant to minor spatial variations and better generalize across various scanning conditions and acquisition protocols.  

The cross-encoder is trained for 50 epochs using the Adam optimizer with an initial learning rate of~$10^{-3}$. The learning rate is adaptively reduced based on validation performance to ensure stable convergence. The learning rate scheduler was chosen as ReduceLROnPlateau, wherein the learning rate was reduced by a factor of 10 when the validation performance stagnated for five consecutive epochs. 
The batch size is set as 64. The key hyperparameters are set as follows: $\lambda_1 = 0.125$, $\lambda_2 = 0.0125$, contrastive learning temperature $\tau=2$, bias $b= -10$, and latent vector size $\rvz =1024$, with the first 25\% (256 dimensions) designated as dynamic features. 

For the classification tasks, we use a multi-layer perceptron with one hidden layer of size 64. Two training settings are explored: (1) Frozen, where the encoder is fixed and only the classifier is trained, and (2) Fine-tune, where both the encoder and classifier are trained jointly. The models are trained using binary cross-entropy loss with class weighting ( pos\_weight $ =1.5$ for AD datasets and 3 for the PPMI dataset) to mitigate class imbalance, treating the NC class as the positive class. The classification training uses the Adam optimizer with an initial learning rate of 0.001, weight decay of 0.0001, and a CosineAnnealingLR scheduler over 50 epochs with a batch size of 64.

The code for the proposed method will be made publicly available.

\subsubsection{Evaluation metrics}
Due to class imbalance, balanced accuracy (BACC) is adopted as the evaluation metric for ADNI and PPMI. Accuracy is used for OASIS, as the dataset is class-balanced. Mean values and standard deviations of BACC are computed over five random splits of training, validation, and test sets.

\subsubsection{Baseline methods}
We compare the proposed method with a 3D CNN-based model trained on Sports-1 Million dataset, which contains 1.1 million YouTube videos, and six self-supervised learning~(SSL) methods: three auto-encoder-based methods -- AE~\cite{AE}, VAE~\cite{VAE} and MAE~\cite{he2022MAE}, two contrastive learning~(CL)-based methods -- SimCLR~\cite{chen2020simple} and SigCLR~\cite{ccaugatan2024sigclr}, and one SSL method that leverages longitudinal data -- LNE~\cite{ouyang2021self}. Five of these SSL methods are based on 3D CNN, namely AE, VAE, SimCLR, SigCLR, and LNE. 
For fairness, these methods use the same encoder and decoder as the cross-encoder (SimCLR and SigCLR do not need a decoder). For VAE, the encoder architecture is changed slightly since it needs to generate both the mean and log variance of the latent variable $\rvz$ instead of directly generating feature representations. More specifically, the encoder has two additional layers, each incorporating two extra Conv3d layers (3×3×3 kernel size each) for generating the mean and log variance. For SimCLR, it has a projection head following the encoder, which contains a linear layer to reduce the dimension of the latent representation, a ReLu activation function, and another linear layer. The temperature parameter is set as 0.5. Unlike the CNN-based approaches, MAE uses a 12-layer vision transformer as the encoder. 

\subsection{Overall Classification Performance and Interpretability}
We first evaluate the classification performance of the proposed method against all baseline approaches for two key tasks: Alzheimer’s disease classification (i.e., NC vs. AD) and progressive mild cognitive impairment classification (i.e., sMCI vs. pMCI). The results, presented in Table~\ref{tab:classification_result}, show that the proposed cross-encoder consistently outperforms all baselines. Specifically, it improves balanced accuracy by 3.20\% and 1.48\% in the two tasks compared with the best-performing baseline, and by 23.68\% and 20.38\% compared with the lowest-performing baseline. These gains are particularly noteworthy given the clinical challenge of distinguishing progressive from stable MCI, where the differences in brain patterns are often subtle and easily overlooked by conventional methods.
\begin{table}[!t]
\begin{center}
\caption{Classification performance on two ADNI tasks, NC vs. AD and sMCI vs. pMCI. Balanced accuracy (mean~$\pm$~standard deviation) is reported, with the best results highlighted in bold.}
\setlength\tabcolsep{4pt}  
\begin{tabular}{lcccc}
\toprule
Method & \multicolumn{2}{c}{NC vs. AD} & \multicolumn{2}{c}{sMCI vs. pMCI} \\
\cmidrule(r){2-3} \cmidrule(r){4-5}
& Frozen & Fine-tune & Frozen & Fine-tune \\
\midrule
C3D~\cite{tran2015learningC3D} & 
71.96$\pm$4.36 & 79.45$\pm$5.13 & 
61.48$\pm$5.22& 68.45$\pm$4.29 \\
AE~\cite{AE} & 
72.52$\pm$3.32 & 78.01$\pm$2.38 &  
60.98$\pm$5.02& 63.25$\pm$3.56 \\
VAE~\cite{VAE} & 
64.98$\pm$2.15 & 75.03$\pm$1.61 & 
54.79$\pm$1.91 & 59.53$\pm$2.69\\
MAE~\cite{he2022MAE} & 
59.49$\pm$4.29 & 61.09$\pm$5.35 &  
50.01$\pm$3.89 & 52.11$\pm$4.51 \\
SimCLR~\cite{chen2020simple} & 
73.58$\pm$1.63 & 82.14$\pm$1.48 & 
62.15$\pm$3.61 & 67.59$\pm$1.73\\
SigCLR~\cite{ccaugatan2024sigclr} & 74.32$\pm$3.06 & 81.37$\pm$2.17 &  
62.50$\pm$4.58&  66.50$\pm$2.10\\
LNE~\cite{ouyang2021self} & 73.36$\pm$3.98 & 81.57$\pm$1.06 & 70.18$\pm$4.56 & 71.01$\pm$4.05 \\
Cross-Encoder (Ours) & \textbf{77.92$\pm$1.92} & \textbf{84.77$\pm$1.08}  & \textbf{70.23$\pm$4.24} & \textbf{72.49$\pm$2.80} \\
\bottomrule
\end{tabular}
\end{center}
\label{tab:classification_result}
\end{table}

\begin{figure}[!t]
\centering
\includegraphics[width=.9\linewidth]{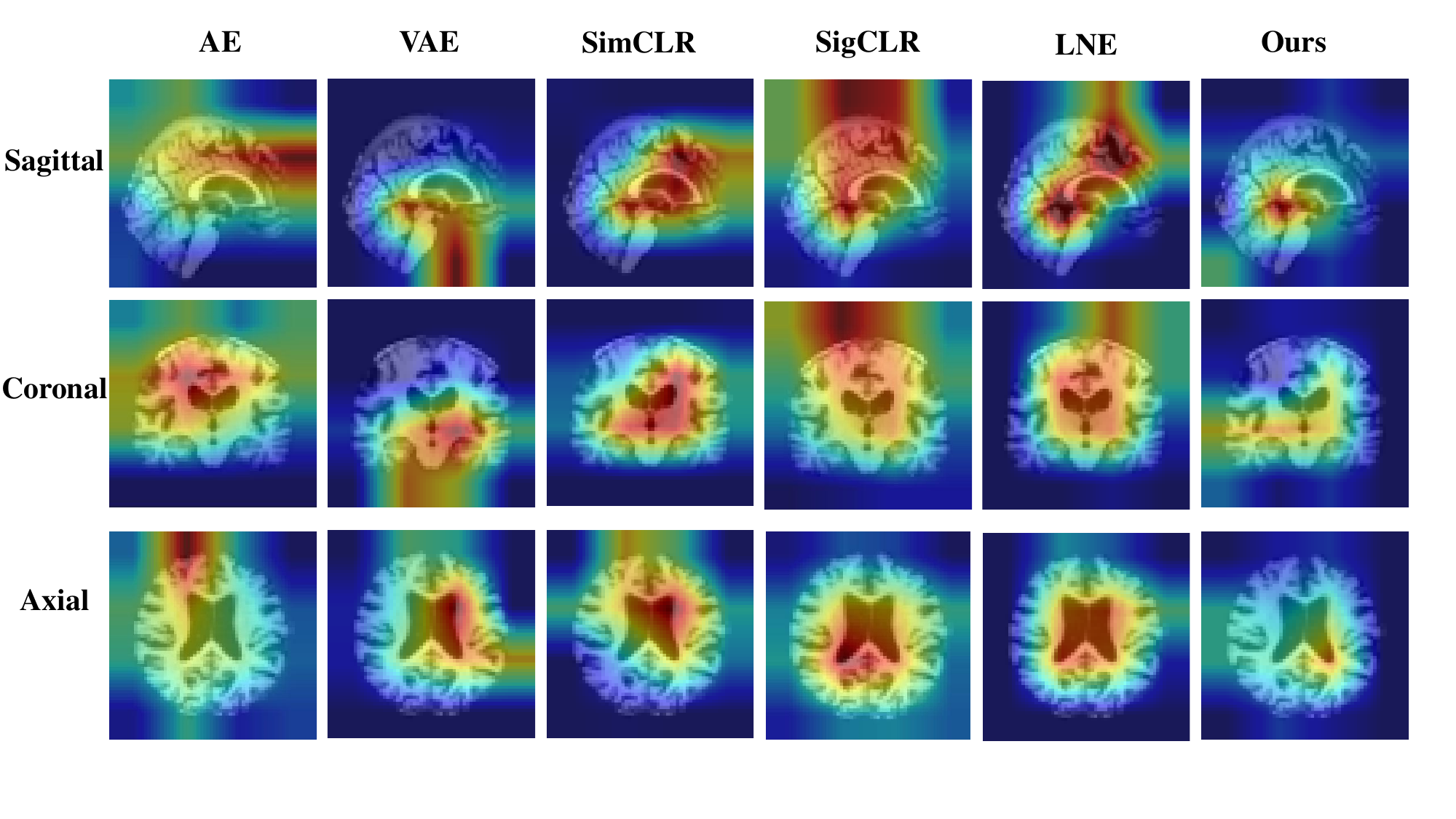}
\caption{Evaluation of interpretability via Grad-CAM. Warmer colors indicate regions with higher discriminative importance. Compared with other methods, our approach yields more compact and localized features that correspond closely to brain regions known to be associated with AD.}
\label{fig:six grad-cam of MRIs}
\end{figure}

Next, we assess interpretability using Grad-CAM~\cite{selvaraju2017gradCam}, a technique that highlights the regions of the input most influential in the classification decision. The analysis is conducted on the fine-tuned model to reflect decision-making behavior after task-specific training. Figure~\ref{fig:six grad-cam of MRIs} shows three MRI slices in the sagittal, coronal, and axial views, where warmer colors indicate higher importance. Existing CNN-based SSL methods (leftmost five columns) tend to produce scattered activation maps without clear and compact key regions. In contrast, our method (rightmost column) focuses on smaller, clinically meaningful regions -- particularly the hippocampus and temporal lobe, both strongly associated with AD~\cite{braak1991neuropathologica}. The targeted focus demonstrates the benefit of incorporating input-gradient regularization, which encourages the extraction of compact, task-relevant dynamic features, thereby enhancing both interpretability and clinical trust.  

\begin{figure}[!t]
\centering
\includegraphics[width=0.8\linewidth]{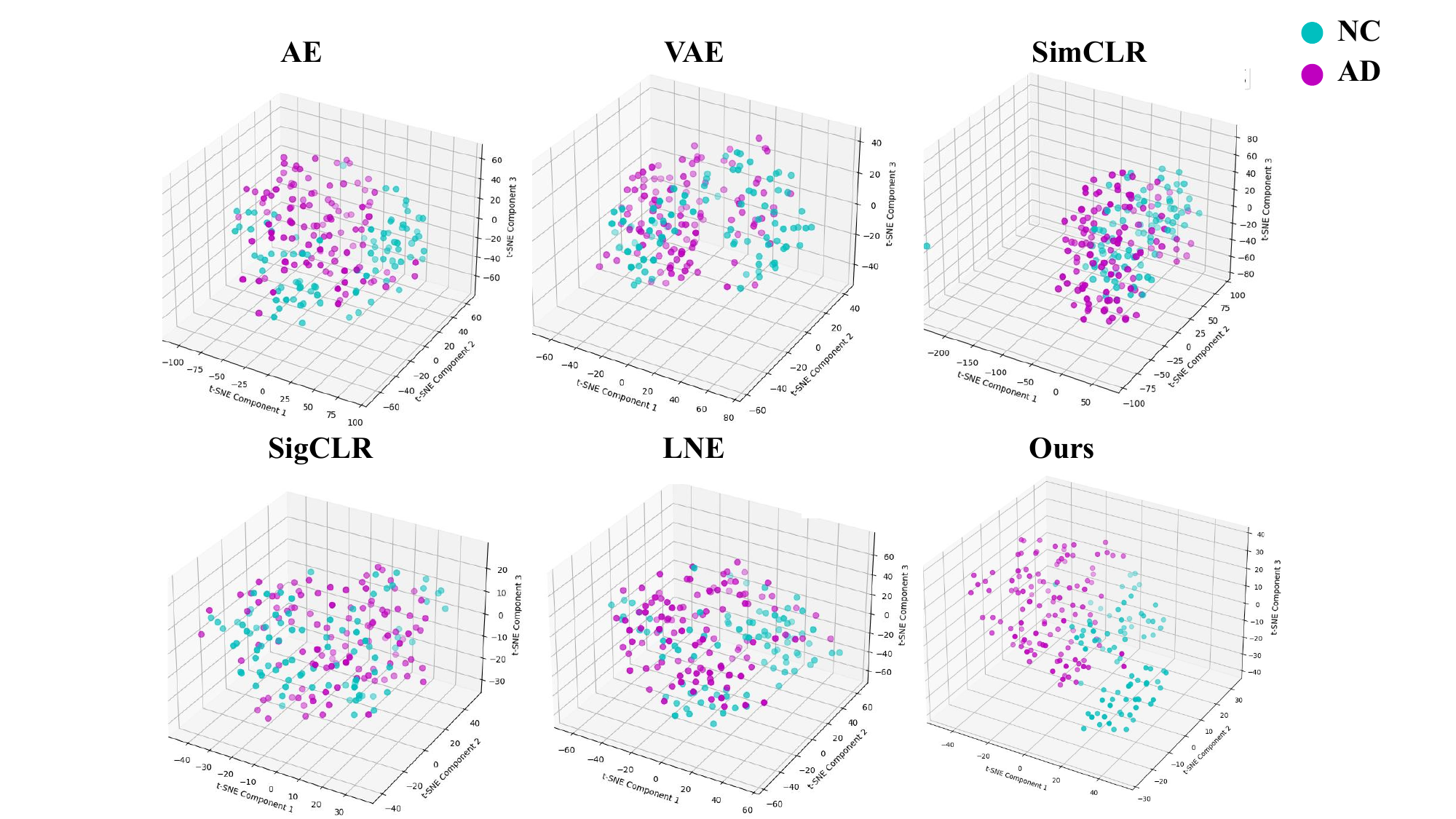}
\caption{t-SNE visualization of latent feature representations. Compared to other methods, our model forms clear and well-separated clusters in three-dimensional space, indicating its ability to learn highly discriminative features that effectively capture the key differences between classes.}
\label{fig:t-SNE}
\end{figure}

To further investigate the discriminative ability of the latent feature representations, t-SNE~\cite{van2008visualizing} is adopted to visualize the fine-tuned representations of the SSL methods in 3D space, with each point colored by its class label. For our method, t-SNE is applied to the dynamic feature $\rvd$, as it directly contributes to classification. For the other methods, t-SNE is applied to their respective latent representations $\rvz$. As shown in Figure~\ref{fig:t-SNE}, after dimensionality reduction, the representations of our method form well-separated clusters, suggesting that the features learned by our model are more discriminative and better capture the underlying class structure.

\subsection{Generalizability of Learned Representations}
\subsubsection{Zero-Shot Classification on OASIS}
To further evaluate the generalization capability of the learned representations, we conduct a zero-shot experiment by directly applying the encoder and classifier fine-tuned on the ADNI dataset to the OASIS-2 dataset for NC vs. AD classification. Neither the encoder is further fine-tuned nor a classifier trained on OASIS-2. This setup evaluates whether the proposed method can maintain decent performance under shifts in data distributions, acquisition protocols, and potential demographic variations, all of which are common in real-world clinical deployment.

As shown in Table~\ref{tab:zero shot result}, cross-encoder achieves the highest accuracy among all baselines and exhibits the smallest performance degradation (only 6.02\% accuracy drop) when transitioning from ADNI to OASIS-2. In contrast, all baseline methods experience an accuracy drop of at least 10.00\%. The results indicate that the proposed method captures disease-related features that transfer well across datasets, likely due to its ability to disentangle compact dynamic representations from MRI scans and reduce overfitting to dataset-specific characteristics.  Furthermore, 
all methods achieve better-than-random performance on OASIS-2, suggesting that the pre-training combined with fine-tuning on ADNI enables meaningful representation learning for NC vs. AD classification. 

\begin{table}[t]
\centering
\caption{Zero-shot classification performance on the OASIS-2 dataset (NC vs. AD) using models trained on ADNI. As the dataset is class-balanced, accuracy (mean~$\pm$~standard deviation) is reported. Performance drop relative to ADNI is indicated in subscript.}
\centering
\vspace{1\baselineskip}
\setlength\tabcolsep{3pt}
\label{tab:zero shot result}
\begin{tabular}{lr}
\toprule
Method & ACC \\
\midrule
C3D~\cite{tran2015learningC3D} &  66.25$\pm$5.90 {\scriptsize\textbf{$\downarrow$13.20\%}} \\
AE~\cite{AE} &  58.75$\pm$6.11 {\scriptsize\textbf{$\downarrow$19.26\%}} \\
VAE~\cite{VAE} & 59.38$\pm$3.20 {\scriptsize\textbf{$\downarrow$15.56\%}}\\
MAE~\cite{he2022MAE} &  50.63$\pm$7.16 {\scriptsize\textbf{$\downarrow$10.46\%}} \\
SimCLR~\cite{chen2020simple} &  68.75$\pm$3.91 {\scriptsize\textbf{$\downarrow$13.39\%}} \\
SigCLR~\cite{ccaugatan2024sigclr} & 70.63$\pm$4.01 
 {\scriptsize\textbf{$\downarrow$10.74\%}}\\
LNE~\cite{ouyang2021self} & 71.25$\pm$5.38 {\scriptsize\textbf{$\downarrow$10.32\%}} \\
Cross-Encoder (Ours) & \textbf{78.75$\pm$3.95} {\scriptsize\textbf{$\downarrow$6.02\%}} \\
\bottomrule
\end{tabular}
\end{table}

\begin{table}[t]
\centering
\caption{Classification performance on PPMI, measured via balanced accuracy~(mean value $\pm$ standard deviation). }
\vspace{1\baselineskip}
\setlength\tabcolsep{3pt}
\label{tab:classifcation result of PPMI}
\begin{tabular}{lrr}
\toprule
Method & Frozen & Fine-tune \\
\midrule
C3D~\cite{tran2015learningC3D} & 55.10$\pm$5.05 & 59.31$\pm$6.09 \\
AE~\cite{AE} & 44.09$\pm$1.72 & 55.60$\pm$2.10 \\
VAE~\cite{VAE} & 51.08$\pm$1.94 & 52.69$\pm$1.96 \\
MAE~\cite{he2022MAE} & 49.08$\pm$2.00 & 52.48$\pm$2.23 \\
SimCLR~\cite{chen2020simple} & 53.48$\pm$1.42 & 62.15$\pm$3.01 \\
SigCLR~\cite{ccaugatan2024sigclr} & 54.84$\pm$3.45 & 62.57$\pm$2.56\\
LNE~\cite{ouyang2021self} & 54.21$\pm$2.12 & 56.15$\pm$1.19 \\
Cross-Encoder (Ours) & \textbf{60.61$\pm$1.82} & \textbf{64.48$\pm$1.80} \\
\bottomrule
\end{tabular}
\end{table}
\subsubsection{Cross-Task Generalization}

To evaluate the model's ability to generalize across different neurodegenerative disorders, we test the classification (PD vs. NC) on the PPMI dataset, and the results are listed in Table~\ref{tab:classifcation result of PPMI}. 

First, due to the dataset shift and task difference between ADNI (Alzheimer) and PPMI (Parkinson), directly applying the encoder trained on ADNI to PD classification yields relatively low balanced accuracy across all methods. However, cross-encoder consistently outperforms all baseline approaches, indicating stronger robustness to domain shifts. Moreover, the performance degradation is less severe for our method compared to other baselines, suggesting superior generalizability of the learned representations.

Second, fine-tuning on the PPMI dataset improves classification performance for all methods. Notably, our method achieves the highest accuracy and demonstrates the largest gain after fine-tuning, underscoring its adaptability to new domains. However, the overall improvement remains less pronounced than that observed on ADNI. This likely reflects two factors: the pre-training on ADNI, which primarily captures AD-specific features, may not fully transfer to PD-related patterns, and the smaller sample size of PPMI limits the effectiveness of fine-tuning.

These findings highlight the challenges of cross-disease generalization in neuroimaging and underscore the value of robust pre-training strategies that can capture shared and transferable features across related neurological conditions. 

\subsection{Ablation Studies}


We conduct ablation studies to investigate the contributions of two key components in the proposed cross-encoder framework: the contrastive learning and the input-gradient regularization. To ensure a fair comparison, all ablation experiments are conducted under the same training and fine-tuning settings as the full model, using the ADNI dataset for pre-training and the same downstream evaluation protocol. Only the specified loss components are modified, while all other architectural and optimization parameters remain unchanged. More specifically, we compare our method with the following three ablation baselines.

\begin{itemize}
    \item w/o $\mathcal{L_\text{CL}}$ + $\mathcal{L_\text{Grad}}$: disables both the proposed contrastive loss $\mathcal{L_\text{CL}}$ and input-gradient regularization $\mathcal{L_\text{Grad}}$, relying solely on the reconstruction loss $\mathcal{L_\text{Rec}}$ for self-supervised learning.
    \item w/o $\mathcal{L_\text{CL}}$: disables the contrastive loss while retaining $\mathcal{L_\text{Grad}}$ and $\mathcal{L_\text{Rec}}$, to assess the role of contrastive alignment in learning informative static features and enhancing the discriminativeness of dynamic features.
    \item w/o $\mathcal{L_\text{Grad}}$: removes the input-gradient regularization while keeping $\mathcal{L_\text{CL}}$ and $\mathcal{L_\text{Rec}}$, to evaluate its effect on the compactness and interpretability of the learned dynamic features.
\end{itemize}

Table~\ref{tab:Ablation Study} summarizes the ablation results, demonstrating the significance of each component within the proposed framework. Removing both the contrastive loss and the input-gradient regularization (w/o $\mathcal{L_\text{CL}}$ + $\mathcal{L_\text{Grad}}$) leads to the largest drop in classification performance, suggesting that relying solely on reconstruction loss is insufficient for learning effective representations. Excluding the contrastive loss (w/o $\mathcal{L_\text{CL}}$) results in a notable decline in balanced accuracy, highlighting that improving the informativeness of static features can aid in enhancing the discriminability of dynamic features. Similarly, the removal of input-gradient regularization (w/o $\mathcal{L_\text{Grad}}$) compromises the compactness of the dynamic features, which is reflected by a decrease in performance. These findings confirm that both contrastive learning and input-gradient regularization are essential for achieving discriminative and interpretable feature representations in our model.

\begin{table}[t]
\centering
\caption{Ablation study of the proposed framework on the ADNI dataset for the NC vs. AD classification task. }
\label{tab:Ablation Study}
\begin{tabular}{lllrr}
\toprule
Method& $\mathcal{L_\text{CL}}$ & $\mathcal{L}_\text{Grad}$  & Frozen & Fine-tune \\
\midrule
w/o $\mathcal{L_\text{CL}}$ + $\mathcal{L_\text{Grad}}$&  &  &  68.95$\pm$1.47 & 79.22$\pm$1.58 \\
w/o $\mathcal{L_\text{CL}}$&  &  \checkmark &  72.40$\pm$1.64 & 82.31$\pm$1.76 \\
w/o $\mathcal{L_\text{Grad}}$& \checkmark & &  74.83$\pm$1.37 & 81.56$\pm$2.87\\
Full model& \checkmark &  \checkmark & \textbf{77.92$\pm$1.92} & \textbf{84.77$\pm$1.08} \\
\bottomrule
\end{tabular}
\end{table}

\begin{figure}[!t]
\centering
\includegraphics[width=0.8\columnwidth]{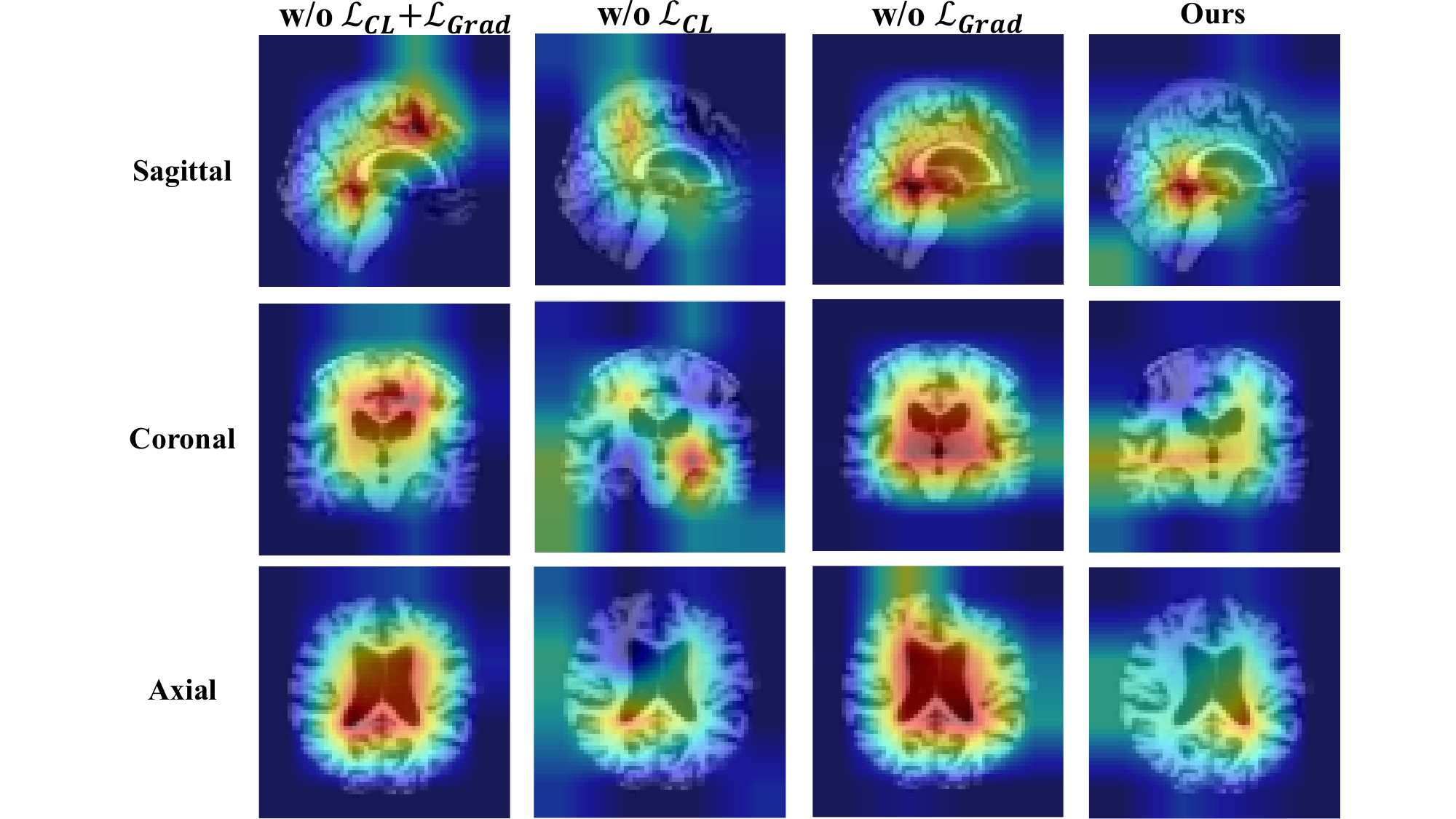}
\caption{Grad-CAM visualization of ablation study. Incorporating either contrastive learning loss or input-gradient regularization improves the compactness of discriminative features, while combining both leads to more focused and clinically relevant activation regions.}
\label{fig: grad-cam of ablation study}
\end{figure}

Figure~\ref{fig: grad-cam of ablation study} presents the Grad-CAM visualizations from our ablation study, illustrating how each component influences both feature learning and interpretability. When both components are removed (i.e., w/o $\mathcal{L_\text{CL}}$+$\mathcal{L_\text{Grad}}$), activations are scattered across the brain, suggesting that the model fails to capture either well-separated feature representations or spatially focused disease-related regions. When only the contrastive learning component is removed (i.e., w/o $\mathcal{L_\text{CL}}$), the high-activation regions (shown in red) become smaller, reflecting the effect of input-gradient regularization that forces the model to focus on a subset of voxels for dynamic features. However, some attention is also misallocated to irrelevant background regions.
Removing only the input-gradient regularization (i.e., w/o $\mathcal{L_\text{Grad}}$) retains some localization in disease-related regions, thanks to the preserved ability to learn meaningful static and dynamic features. However, the highlighted regions remain scattered due to the absence of input-gradient regularization. The full model (rightmost column) produces more compact and disease-specific activation regions, most prominently in the hippocampus and temporal lobe, which align with established neuropathological findings.

These findings underscore the necessity of the proposed both components: contrastive learning for disentangling static and dynamic features, and input-gradient regularization for enforcing spatial sparseness in attention.

\subsection{Sensitivity Analysis}

\begin{table}[t]
\caption{Classification performance of the proposed method for different sizes of the dynamic feature ($|\mathbf{d}|$). Static feature ($\rvs$) result is included as a separate reference.}
\label{tab:Ablation of d}
\centering
\begin{tabular}{lrr}
\toprule
$|\rvd|$ & Frozen & Fine-tune \\
\midrule
128 & 66.70$\pm$5.15 & 73.92$\pm$2.21 \\
256 & \textbf{77.92$\pm$1.92} & \textbf{84.77$\pm$1.08} \\
512 & 74.39$\pm$2.79 & 81.41$\pm$1.36 \\
\midrule
$|\rvs|=768$ & 67.55$\pm$4.17 &{73.19$\pm$2.26} \\
$[|\rvd|,|\rvs|]=1024$ & 74.03$\pm$3.59 & 81.85$\pm$2.83 \\
\bottomrule
\end{tabular}
\end{table}

We further investigate the impact of the dimensionality of the dynamic feature $\rvd$ on classification performance. Throughout these experiments, the total latent space dimensionality is fixed at 1024. As shown in Table~\ref{tab:Ablation of d}, increasing the dimensionality of~$\rvd$ (denoted as $|\mathbf{d}|$) from 128 to 512 initially improves performance but eventually leads to a decline. This trend highlights the importance of balancing static and dynamic features. The static features should remain the dominant representation, providing stable, context-rich information, while the dynamic features serve as a smaller, targeted subset that captures discriminative, clinically relevant variations. When $|\rvd|$ is too small, the model lacks sufficient capacity to encode meaningful dynamic patterns, reducing discriminative ability. However, when $|\rvd|$ becomes too large, e.g., $|\rvd|$=512 (half of the latent vector), the performance also drops. This is because the dynamic feature begins to contain redundant information, which dilutes its discriminative ability. Based on these findings, we set $|\mathbf{d}| = 256$ for all experiments.

We also compare classification performance when using only the dynamic feature $\mathbf{d}$, only the static feature $\mathbf{s}$, or their combination $\mathbf{d} + \mathbf{s}$. As shown in Table~\ref{tab:Ablation of d}, using only $\mathbf{d}$ substantially outperforms both $\mathbf{s}$ alone and the combined features, suggesting that the dynamic feature captures the most discriminative information relevant to disease status. This result further supports the effectiveness of disentangling dynamic features from static ones, ensuring that disease-related variations are preserved without being overshadowed by non-discriminative content.

\section{Conclusion}

In this work, we propose a self-supervised cross-encoder framework that effectively leverages longitudinal MRI data to learn disentangled feature representations, explicitly separating static and dynamic brain features. Additionally, we incorporate contrastive learning and input-gradient regularization to further enhance feature learning. Extensive experiments demonstrate that our method achieves superior performance in downstream classification tasks across both within-dataset and cross-dataset settings, as well as improving interpretability.

Despite these promising results, several limitations remain. First, the current classification framework does not utilize valuable complementary information such as age, medical history, and other clinical data, which could potentially enhance performance. Second, the generalization ability of the model across different but related classification tasks is still limited, posing challenges for broader clinical application. Future work will focus on integrating multimodal data and improving cross-task generalization to further strengthen the clinical utility of the proposed framework.

\bibliographystyle{ieeetr}
\bibliography{refs}

\end{document}